\def\year{2017}\relax
\documentclass[letterpaper]{article}
\usepackage{aaai17}
\usepackage{times}
\usepackage{helvet}
\usepackage{courier}
\usepackage{amsmath}
\usepackage{dcolumn}
\usepackage{hhline}
\usepackage{verbatim}
\usepackage{multirow}
\usepackage{array}
\usepackage{graphicx}

\newcolumntype{L}[1]{>{\raggedright\let\newline\\\arraybackslash\hspace{0pt}}m{#1}}
\newcolumntype{C}[1]{>{\centering\let\newline\\\arraybackslash\hspace{0pt}}m{#1}}
\newcolumntype{R}[1]{>{\raggedleft\let\newline\\\arraybackslash\hspace{0pt}}m{#1}}

\title{Efficient Transfer Learning Schemes for Personalized Language Modeling \\
using  Recurrent Neural Network}

\author{Seunghyun Yoon\textsuperscript{1,2},  Hyeongu Yun\textsuperscript{1},  Yuna Kim\textsuperscript{2},  Gyu-tae Park\textsuperscript{2} \and Kyomin Jung\textsuperscript{1,3}\\
\textsuperscript{1}Department of Electrical and Computer Engineering, Seoul National University, Seoul, Korea\\
\textsuperscript{2}Software R\&D Center, Samsung Electronics Co., Ltd., Seoul, Korea\\
\textsuperscript{3}Automation and Systems Research Institute, Seoul National University, Seoul, Korea\\
\{mysmilesh, kjung\}@snu.ac.kr, hyeongu.yun.1989@gmail.com, \{yuna1.kim, gyutae.park\}@samsung.com
}

\begin{document}

\maketitle

\begin{abstract}
In this paper, we propose an efficient transfer leaning methods for training a personalized language model using a recurrent neural network with long short-term memory architecture. With our proposed fast transfer learning schemes, a general language model is updated to a personalized language model with a small amount of user data and a limited computing resource. These methods are especially useful for a mobile device environment while the data is prevented from transferring out of the device for privacy purposes. Through experiments on dialogue data in a drama, it is verified that our transfer learning methods have successfully generated the personalized language model, whose output is more similar to the personal language style in both qualitative and quantitative aspects.
\end{abstract}

\section{Introduction}
\label{intro}
Recently there has been a considerable interest in language modeling due to various academic and commercial demands. Academically, many studies have investigated this domain such as machine translation, chat-bot, message generation, image tagging and other language-related areas. Commercially, it can be used as a core technology for providing a new application on consumer products or services. For instance, an automatic message-reply prediction service can be launched in mobile devices, thus helping a user to send a reply message when he/she is not provided with a proper input interface.

To model the language of human dialogue, a recurrent neural network (RNN) structure is known to show the state of the arts performance with its ability to learn a sequential pattern of the data \cite{sutskever2014sequence}. Among the RNN structures, a Long Short-Term Memory RNN (LSTM-RNN) and its variants are successfully used for language modeling tasks \cite{hochreiter1997long,cho2014learning}. However, as a kind of deep learning technique, the LSTM-RNN and the RNN structure requires both a large number of data and huge computing power to train the model properly. Hence any attempts for applying the RNN structure to personalized language modeling are mainly constrained by the following two limitations. First, personal mobile devices contain private message data among close acquaintances, so users seldom agree to transfer their log out of the devices. This causes a limitation of gathering the whole user data to common computing spaces, where high-performance machines are available. Second, in relatively small computing machines, i.e., smart phone, it is not always-guaranteed to have enough resources to train a deep model within the devices.

To resolve these limitations, we propose fast transfer learning schemes. It trains a base model with a large dataset and copies its first n-many layers to the first n-many layers of a target model. Then the target model is fine-tuned with relatively small target data. Several learning schemes such as freezing a certain layer or adding a surplus layer are proposed for achieving the result. In experiments, we trained a general language model with huge corpus such as an Workshop on Statistical Machine Translation (WMT) data\footnote{Available from  ``http://www.statmt.org/wmt14/translation-task.html\#download/"} and a movie script data by using powerful computing machines, and then transferred the model to target environment for updating to be a personalized language model. With this approach, the final model can mimic target user's language style with proper syntax.

In the experiments, we trained the general language model with literary-style data and applied the transfer learning with spoken-style data. Then we evaluated the model output for sentence completion task in a qualitative and a quantitative manner. The test result showed that the model learned the style of the target language properly. Another test was conducted by training the general language model with the script of the drama, ``Friends," and by applying transfer learning with main character corpora from the script to generate the personalized language model. The message-reply prediction task was evaluated with this model. The test result shows higher similarity between the output of the personalized language model and the same user dialogue than the one between the output of the personalized language model and other users' dialogues.

The contributions of this paper are as follows. First, we propose efficient transfer learning schemes for personalized language modeling, which is the first research on transfer learning for RNN based language models with privacy preserving. Second, we show the applicability of our research to the target scenario in the short message reply application by training the model in the similar environment to that of the mobile device, and highlight its test results.

%The remainder of the paper is organized as follows. Section~\ref{model} describes the two language model, sentence completion and message-reply prediction, and the proposed transfer learning schemes for the model personalization. In Section~\ref{measures}, a novel similarity measure is introduced which is designed to evaluate the language model output. The datasets used for experiments are described in Section~\ref{datasets}, and the experimental results are presented in Section~\ref{experiments}. The previous works related to the language modeling and the transfer learning are summarized in Section~\ref{related work}. Finally, Section~\ref{conclusion} offers some plans for future work.

\section{Architecture for Personalized Language Model}
\label{model}

As we are focusing on a personalized language modeling with the preservation of user data, we generate two types of language models. First is a sentence completion language model, which can complete sentences with a given n-many sequence of words. Second is a message-reply prediction language model, which can generate a response sentence for a given message. The output of both models implies user characteristics such as preferable vocabulary, sentence length, and other language-related patterns.

To achieve this result, we trained the language model with a large amount of general data in powerful computing environments, and then applied the transfer learning in relatively small computing environments. We assume that this method would be applied to mobile devices. As we are taking the preservation of privacy into consideration, the transferred model is retrained within the local environments such as mobile devices, and no personal data is sent out of the devices. This could have been accomplished using the proposed transfer learning schemes in RNN-LSTM architecture.

\subsection{Sentence Completion Language Model}

A sentence completion model completes a sentence with the given word sequence $\boldsymbol{X}= \{x_1,x_2, \dots, x_T\}$, where $x_N$ is a word ($N=1, 2, \dots, T$). The model can predict the next word $x_{N+1}$ with given word sequence $x_{1:N}$. By repeating the prediction until the output word reaches the end-of-sentence signal, ``$<eos>$," the whole sentence can be generated.

% with the given model parameter $\Theta$ and input word sequence in time step $T$. The $\Theta$ includes weights and biases of the embedding layer, the deep LSTM layers, and the softmax output layer.

The model is similar to that of \cite{graves2013generating}, and we put the 1,000-dimension word-embedding layer right after the input layer. Then 3 deep LSTM layers with 100 LSTM cells each and without peephole connection are used for learning the sequential pattern of the sentences. 
% The output of the word embedding layer can be calculated by $O_{embed}=X_{input}\times W_{embed}$, where $X_{input}$ is one-hot-encoded input and $W_{embed}$ is the embedding matrix. 

The output probability to the input sequence $\boldsymbol{X}$ and the training objective are 
\begin{equation}
\begin{aligned}
% & p(y_1,\dots,y_{T'}|x_1,\dots,x_T)=\prod_{t=1}^{T'}p(y_t|y_1,\dots,y_{t-1}) \\
% & 1/
& p(\boldsymbol{Y|X})=\prod_{t=1}^{T}p(y_t|x_{1:t-1}) \\
& \textit{L}= -\dfrac{1}{|T|}\sum\limits_{t=1}^{T} x_{t+1}\log p(y_t|x_{1:t-1}),
\end{aligned}
\end{equation}
where $X$ is a word sequence in the sentence, $Y$ is a model output sequence $\boldsymbol{Y}=\{y_1,y_2, \dots, y_{T}\}$

\subsection{Message-Reply Prediction Language Model}

A message-reply prediction model generates a response sentence for a given message. It is similar to the sentence completion language model except that the message sentence is encoded and used as a context information when the model generates a response word sequence. Our approach is inspired by the sequence-to-sequence learning research \cite{sutskever2014sequence} that is successfully applied to a machine translation task. The message word sequence $\boldsymbol{X}=\{x_1, x_2, \dots, x_T\}$ is fed into the model, and the last hidden state is used as context information $c_T$. With this context information, the next sequence word is predicted similarly to that in the sentence completion language model case. During implementation, we used 1,000-dimension word embedding and 3-deep LSTM layers with 100 LSTM cells in each layer. The output probability and the training objective are

\begin{equation}
\begin{aligned}
& p(\boldsymbol{Y|X})=\prod_{t=1}^{T'}p(y_t|c_T,y_{1:t-1})\\
& L = -\dfrac{1}{|T'|}|\sum\limits_{t=1}^{T'} z_t\log p(y_t|c_T, y_{1:t-1}),
\end{aligned}
\end{equation}
where $X$ is a word sequence in the message sentence, $Z$ is a target word sequence in the response sentence $\boldsymbol{Z} = \{z_1,z_2, \dots, z_{T'}\}$, $Y$ is a model output sequence $\boldsymbol{Y}=\{y_1,y_2, \dots, y_{T'}\}$, $c_T$ is the encoding vector for the message sentence.

\subsection{Fast Transfer Learning Schemes}
To generate a personalized language model with a small amount of user data and limited computing resources, transfer learning is essential. In the private data preservation scenario, we investigate three fast transfer learning schemes. Each scheme is described below:
\begin{itemize}
\item Scheme 1, relearn the whole layer: As a baseline, we retrain the whole model with private data only and compare the result with the two other schemes below. Because of the retraining of the LSTM layers in their entirety, this scheme requires more computing power than the other two schemes.
\item Scheme 2, surplus layer: After the training of the model with general data, a surplus layer is inserted between the output layer and the last of the deep LSTM layers. Then, with private data, we update only the parameters of the surplus layer in the transfer learning phase. We assume that a user's parlance could be modeled by learning additional features in the user's private data.
\item Scheme 3, fixed first n layers: After training the model with general data, we fix the parameters in the first n LSTM layers (layer 1 and layer 2 in our experiments) and train remaining parameters in the transfer learning phase. We assume that the user's parlance is a subset of the general pattern and the last layer plays the key role in determining this pattern.
\end{itemize}

\begin{table*}[ht]
\small
\centering
\begin{tabular}{ |C{0.23\columnwidth}|C{0.23\columnwidth}|C{0.23\columnwidth}|C{0.23\columnwidth}|C{0.23\columnwidth}|C{0.23\columnwidth}|C{0.23\columnwidth}| }

	\hline
%     \abovespace\belowspace
		&	character\_1	&	character\_2	&	character\_3	&	character\_4	&	character\_5	&	character\_6	\\
    \hhline{|=|=|=|=|=|=|=|}
character\_1	&\textbf{5.5667}		&5.7957		&5.7814		&5.7781		&5.7026		&5.8147	\\
	\hline
character\_2	&6.1770		&\textbf{5.6097}		&5.8256		&5.8254		&5.7397		&5.8562	\\
	\hline
character\_3	&6.1741		&5.8702		&\textbf{5.6057}		&5.8056		&5.7101		&5.8623	\\
	\hline
character\_4	&6.1990		&5.8672		&5.8240		&\textbf{5.5726}		&5.7102		&5.8689	\\
	\hline
character\_5	&6.1948		&5.8592		&5.8176		&5.7898		&\textbf{5.5099}		&5.8460	\\
	\hline
character\_6	&6.1782		&5.8415		&5.8279		&5.8132		&5.7120			&\textbf{5.6171}	\\
	\hline
\end{tabular}
\caption{Quantitative measure result of dialogues among main characters.  Character\_1 to character\_6 are Chandler, Joey, Monica, Phoebe, Rachel, and Ross, respectively. A lower value indicates that the two sets compared have similar distributions and are, thus, similar in style.}
\end{table*}

\section{Measures}
\label{measures}

The perplexity is one of the popular measures for a language model. It measures how well the language model predicts a sample. However, it is not good at measuring how well the output of the
language model matches a target language style. Another measure, the BLEU score algorithm \cite{papineni2002bleu}, has been widely used for the automatic evaluation of the model output. However, it cannot be applied directly to measuring a quality of the personalized model output because it considers the similarity between one language and the target language. Other research was conducted on proving authorship and fraud in literature, for instance, Jane Austen's left-over novel with partially completed \cite{morton1978literary}. This research counted the occurrence of several words in the literature, compared their relative frequencies with those of the words in the target literature, and concluded that the target literature was a forgery. This approach could be applied to a text evaluation where a large amount of data is available and certain words are used more frequently. In spoken language, such as in the message-reply case, however, whole word distribution must be considered instead of considering the occurrence of several words, because the data is usually not enough than the literature case. So, we use a simple and efficient metric to measure the similarity between the user style and the output of the personalized model.

% \subsection{Proposed Method}
An output of a personalized language model can be measured by calculating the cross entropy between the word distribution of the model output and that of the target data. Word distribution can be acquired by normalizing a word histogram which is calculated based on word counts in the target corpus. Equation (3) shows the metric formulation.
\begin{equation}
\begin{aligned}
& Y_1=g( f_{LM}( M_i ) ), Y_2=g( T_i ) \\
& measure = Cross~Entropy(Y_1, Y_2), \\
\end{aligned}
\end{equation}

where $M_i$ is a message $\in{D_{test}}$, $T_i$ is a corpus $\in{D_{target}}$, $f_{LM}$ is a language model, $g(\cdot)$ calculates word distribution with given corpus, CrossEntropy(p, q) is $- \sum_{x} p(x) \log q(x)$.

The characteristics of a user speech can mainly be distinguished by the word dictionary. Thus, this metric tries to measure the differences of the word dictionary among the comparing set. Table 1 shows the quantitative measure results from the dialogue set of the main characters in drama data from ``Friends," a famous American television sitcom. In the figures, ``character\_1" to ``character\_6" are the main characters of the drama (Chandler, Joey, Monica, Phoebe, Rachel, and Ross, respectively). The dialogues were measured against one another by using the cross entropy metric. As shown in the table, the lower cross entropy value among the same character's dialogue was calculated, and the higher value was calculated among the different character's dialogues as expected. This result demonstrates that the cross entropy metric can be used to measure the similarities among the members of the set.

\section{Datasets}
\label{datasets}
\begin{itemize}
\item  \textbf{WMT’14 ENG Corpus:} The WMT'14 dataset includes several corpora. We only use an English part of the $10^9$ French-English corpus. The dataset was crawled data from the bilingual web pages of the international organizations \cite{callison2011findings}. Thus, it contains high quality formal written language data. It consists of 21,000,000 sentences.

\item \textbf{English Bible Corpus:} The English bible corpus is another type of written language data. It is useful data that differs from the WMT'14 dataset not only in the frequent vocabulary type but also in the average sentence length. It consists of 31,102 sentences.
% and 4,100,000 words in total.

\item \textbf{Drama Corpus:} To collect spoken language data, we use drama data from ``Friends" from opensubtitles\footnote{Available from ``http://www.opensubtitles.org/"}. We extracted 69,000 sentences from dialogues, which we used to train a sentence completion language model. For the message-reply prediction language model, pairwise data is required. Among the extracted data, two consecutive sentences of different characters are linked into a single sentence to generate pairwise data.

\item \textbf{Main Character Corpora:} From the drama corpus, we extract main character corpora to model personal users. For example, the Chanlder (one of the main characters in ``Friends") corpus consisted of 8,406 lines and the Rachel (another major character in ``Friends") corpus consisted of 9,194 lines. The former data could represent a male adult, and the latter data could represent a female adult. We assume that those amounts of data could be gathered in a user device for the personalizing language model.

\end{itemize}

\begin{table*}[htb]
\label{literal spoken wmt friend table}
\small
\centering
\begin{tabular}{|C{0.25\columnwidth}|L{1.7\columnwidth}|}
\hline
\multicolumn{2}{|c|}{General language model}                                                                                                                                                                                                \\ \hline
\multicolumn{2}{|c|}{it is possible, however, that investments are only being upgraded in one or more labour conclusions.}                                                                                                                      \\ \hhline{|==|}
\multicolumn{2}{|c|}{Personal language model 1}                                                                                                                                                                                             \\ \hline
scheme 1                       & it is possible, however, we all offered to break this time you have tools.                                                                                                                                 \\ \hline
scheme 2                       & it is possible, however, he’s not bad enough than rachel’s feeling.                                                                                                                                        \\ \hline
scheme 3                       & it is possible, however, they’re right. you can’t wait this.                                                                                                                                               \\ \hhline{|==|}
\multicolumn{2}{|c|}{Personal language model 2}                                                                                                                                                                                             \\ \hline
scheme 1                       & it is possible, however, ye are able to cut off the cross, and remain in the fire, where they likewise eat shall ye be among them; and ye shall fight against your brethren them taken abroad in our lord. \\ \hline
scheme 2                       & it is possible, however, this mountain in the eleventh offering of the doctrine of god; and all the earth shall be there without help.                                                                     \\ \hline
scheme 3					   & it is possible, however, the wilderness shall eat his drink.                                                                                                                                               \\ \hline
\end{tabular}
\caption{Sample model output of general language model and personalized language model. The general language model used WMT'14 data, personalized language model 1 used ``Friends" drama data, and personalized language model 2 used the English bible data. Scheme\_1 to scheme\_3 are relearn whole, surplus layer and fixed-n layer, respectively. The output was generated with the given input sequence, ``It is possible, however"}
\end{table*}

\section{Experiments}
\label{experiments}
We mainly conduct two types of experiments. The first one is a sentence completion experiment, and the other one is a message-reply prediction experiment. In the former case, we train a general language model with literary-style data and apply a proposed transfer learning scheme with spoken-style data to achieve a personalized language model. With this setting, the difference between general and personalized language models can be measured in a quantitative and a qualitative manner. For the latter case, we use dialogue-style data such as drama scripts to train a general language model. From the drama scripts, some characters' data are taken apart and are used to train the personalized language model. With this setting, the output of the personalized model is compared to the original dialogue of the same character.

\subsection{Literary-Style to Spoken-Style Sentence Completion}
We train a general language model of literary-style with the WMT'14 corpus. We then apply a transfer learning scheme with ``Friends" drama data for the model to learn the spoken-style language. Training the general language model took about 10 days then we spent another 4 hours training the personalized language model in each scheme. A ``titan-X GPU" and a ``GeForce GT 730 GPU" were used for these experiments. The latter GPU is one of the low-end GPU series of which computing power was similar to that of latest mobile GPUs such as ``Qualcomm Adreno 530" in ``Samsung Galaxy S7" or ``NVIDIA Tegra K1" in ``Google Nexus 9". For a vocabulary setting, we construct our dictionary as 50,002 words, including ``$<eos>$" to mark ends of sentence and ``**unknown**" to replace unconsidered vocabulary in the data. The out-of-vocabulary rate is about 3.5\%. 

The ``general language model" in Table 2 shows the sample output of the general language model trained with document-style data, and the ``personal language model 1" in Table 2 shows the sample output of the personalized language model trained with human-dialogue-style data. Scheme\_1 to scheme\_3 are relearn-whole, surplus layer, and fixed-n layer, respectively. Given input word sequence for the test was, ``It is possible, however." As can be seen in the table, both outputs differ in length and style. The sentence completed using the general language model tends to be longer than that of obtained using the personalized language model. This result indicates that the personalized language model is properly trained with the spoken language characteristics because human dialogue is usually briefer than the language in official documents.
 
 We also apply the transfer learning schemes with some of the English bible data. The same general language model, which involved previously training with the WMT'14 corpus for 10 days, is used. English bible data is added and employed in training for another 4 hours using proposed transfer learning schemes.
 
The ``personalized language model 2" in Table 2 shows the sample output of the personalized language model trained with another style of document data, English bible data. As shown in Table 2, the output of the personalized language model contains more bible-like vocabulary and sentence styles.

\subsection{General-Style to Personal-Style Message-Reply Prediction}

We simulate the message-reply prediction scenario using the drama corpus. The script of the drama, ``Friends," is used to train a general language model, and two main character corpora are used to generate a personalized language model. For this message-reply prediction experiment, we use a vocabulary size of 18,107, and the out-of-vocabulary rate is about 3.5\%. In the message-reply prediction case, pairwise data is generated by extracting the drama corpus of each character and concatenating two consecutive sentences of different characters to form one single message-reply sentence data. We insert the word ``$<eos>$" between the message and reply to mark the border separating them. This pairwise data is used for the training, and only the message part of the pairwise data is used for the message-reply prediction. During implementation, it took about a day to train the general language model with the ``Friends" corpus and another 4 hours to train the personalized language model with two main character corpora. The ``titan-X GPU" and the ``GeForce GT 730 GPU" was used for these experiments. Validation messages-reply sentences of 1,281 are randomly sampled from the ``Friends" corpus for tracking validation curve and another 753 test messages are prepared for predicting the responses. These data remained unseen from training phase. The word distributions of the model output from the test messages and the target corpus data are calculated to measure their similarity.

\begin{figure}[t]
\small
\centering
\includegraphics[width=0.85\columnwidth]{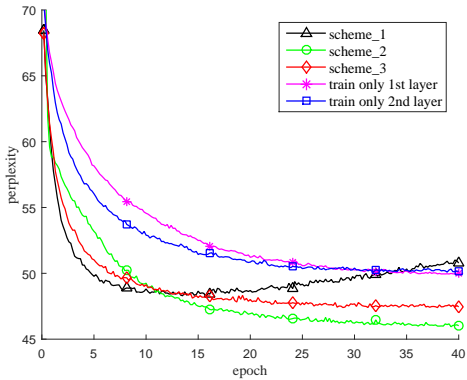}
\caption{Validation curve for each schemes. Scheme\_1 is re-learn whole, scheme\_2 is surplus layer and scheme\_3 is fixed-n layer (train 3rd layer only).}
\label{valid_curve}
\end{figure}

Figure 1 shows the validation curve while training. Perplexity values from various model output are plotted. The perplexity of baseline model, ``scheme\_1", decreases until around epoch 10, and then it starts to increase because model is over-fitted to training data. The proposed ``scheme\_2" and ``scheme\_3", however, show continuous decreasing tendency and reach lower perplexity values compared to that of the baseline model. It is interesting that proposed methods achieve lower perplexity than baseline while saving computing power with reduced parameters.

Table 3 shows the performances of various models measured with the same validation dataset used in Figure 1. An unpruned n-gram language models using modified Kneser-Ney smoothing are used for performance comparisons \cite{chen1996empirical}. The n-gram models were trained by using KenLM software package \cite{heafield2013scalable}. The chandler n-gram model was trained with ``Chandler'' corpus and the friends n-gram model was trained with ``Friends'' corpus. The proposed scheme\_1 to scheme\_3 were trained with ``Chandler'' corpus from ``Friends'' general language model. We see that our proposed schemes outperform the n-gram models (n=3 and 5).

To check the influence of training data size (number of sentences) in personalized language model, we trained the general language model (trained with ``Friends" corpus, message-reply prediction model) with different sizes of personal (``chandler" and ``rachel") dataset. The proposed scheme\_2 method was used for this test. Table 4 shows evaluation results of the trained models. Dataset '0' means the model is not trained with personal dataset. The perplexity shows lower value as we use more dataset in training, and it outperforms ``friends 5-gram'' model from the 2,000 dataset cases.

\begin{table}[t]
\small
\centering
\begin{tabular}{|C{0.5\columnwidth}|C{0.3\columnwidth}|}
\hline
model                 & perplexity    \\ \hhline{|=|=|}
chandler 3-gram       & 77.93         \\ \hline
chandler 5-gram       & 76.85         \\ \hline
friends 3-gram        & 68.55         \\ \hline
friends 5-gram        & 56.69         \\ \hhline{|=|=|}
scheme\_1 (base line) & 48.17		  \\ \hline
scheme\_2             & \textbf{46.02}         \\ \hline
scheme\_3             & 47.45         \\ \hline
\end{tabular}
\caption{Performances of models measured with the same validation dataset used in Figure 1. The chandler n-gram model was trained with ``Chandler'' corpus and the friends n-gram model was trained with ``Friends'' corpus. The scheme\_1 model is over-fitted to training data (see Figure 1), and the lowest value is 48.17.}
\label{measure comparison}
\end{table}

\begin{table}[t]
\small
\centering
\begin{tabular}{|C{0.18\columnwidth}|C{0.10\columnwidth}|C{0.10\columnwidth}|C{0.10\columnwidth}|C{0.10\columnwidth}|C{0.10\columnwidth}|}
\hline
dataset    & 0      & 1000  & 2000  & 4000  & 6000  \\ \hline
perplexity & 68.38 & 58.93 & 52.94 & 48.37 & 47.07 \\ \hline
\end{tabular}
\caption{Performances of models with different number of sentences in training dataset (lower is better). ``Friends'' corpus was used pre-training the general model, and ``Chandler'' and ``Rachel'' corpus was used training the personalized model with the proposed scheme\_2 method. Dataset '0' means the model is not trained with personal dataset.}
\label{number of data}
\end{table}

\begin{table*}[ht]
\label{message response cross entropy chandler and drama}
\small
\centering
\begin{tabular}{|C{0.35\columnwidth}|C{0.3\columnwidth}|C{0.25\columnwidth}|C{0.25\columnwidth}|C{0.25\columnwidth}|C{0.25\columnwidth}|}
\hline
\multirow{2}{*}{target corpus} & \multirow{2}{*}{model scheme} & \multicolumn{4}{c|}{epoch}                         \\ \cline{3-6} 
                               &                         & 0                       & 10     & 20     & 40     \\ \hhline{|=|=|=|=|=|=|}
\multirow{3}{*}{friends}       & 1               & \multirow{3}{*}{6.1222} & 6.5880 & 6.6049 & 6.6900 \\ \cline{2-2} \cline{4-6} 
                               & 2               &                         & 6.8388 & 6.6090 & 6.6312 \\ \cline{2-2} \cline{4-6} 
                               & 3               &                         & 6.6667 & 6.6974 & 6.5097 \\ \hline
\multirow{3}{*}{chandler}      & 1               & \multirow{3}{*}{6.9857} & 6.0496 & 6.0292 & 6.0398 \\ \cline{2-2} \cline{4-6} 
                               & 2               &                         & 6.3374 & 6.0862 & 6.1048 \\ \cline{2-2} \cline{4-6} 
                               & 3               &                         & 6.0419 & 6.0499 & \textbf{5.9429} \\ \hline
\multirow{3}{*}{bible}         & 1               & \multirow{3}{*}{7.8622} & 8.4750 & 8.2865 & 8.4472 \\ \cline{2-2} \cline{4-6} 
                               & 2               &                         & 8.5955 & 8.3145 & 8.3594 \\ \cline{2-2} \cline{4-6} 
                               & 3               &                         & 8.4471 & 8.3803 & 8.3028 \\ \hline
\end{tabular}
\caption{Cross entropy measure between the language model output and the training data corpus, the ``Friends" drama corpus, the``Chandler" corpus and the ``Bible" corpus. Scheme\_1 to scheme\_3 are relearn whole, surplus layer and fixed-n layer, respectively. The ``epoch 0" case means the initial model state trained from general language corpus, ``friends" corpus. Thus cross entropy with ``friends" target corpus shows lower value than that of ``chandler" and ``bible" target corpus cases. The lower value indicates that the language model output is similar in style to the compared target corpus}
\end{table*}

Table 5 indicates the cross entropy measure between the output of ``scheme\_1" to ``scheme\_3" model and that of the target corpus, the ``friends" drama corpus, the ``chandler" corpus, and the ``bible" corpus. It shows the similarity between the personalized model output and the target corpus as the number of training epoch increasing. The general model was pre-trained with the ``Friends'' corpus and the ``Chandler'' corpus was used training personalized model. Each Model is selected from various training epoch (0, 10, 20 and 40) and schemes, and test messages of 753 are used for the reply generation with the selected model used. As the table shows, the cross entropy measure has the highest value when the target corpus is the ``bible'' as expected because it is written in different style than dialogues in drama script. For the drama script case, the cross entropy measured with the ``chandler" corpus shows the lowest value among schemes. This result reveals that the personalized language model is trained properly from the general language model. Thus it is more similar in style to the target data corpus than the general language model. The ``epoch 0" case means the initial model state trained from general language corpus, ``friends" corpus. Thus cross entropy with ``friends" target corpus shows lower value than that of ``chandler" and ``bible" target corpus cases.

% \begin{table}[ht]
% \label{message response sample}
% \small
% \centering
% \begin{tabular}{|L{0.43\columnwidth}|L{0.43\columnwidth}|}
% \hline
% \multicolumn{1}{|c|}{Message}         & \multicolumn{1}{c|}{Response} \\ \hhline{|=|=|}
% more coffee?                          & all right one.                \\ \hline
% Which goes where?                     & I have no idea.               \\ \hline
% is that a dinosaur tie?               & this is nice.                 \\ \hline
% what?                                 & sorry, is that okay?          \\ \hline
% you see that?                         & you’re excited?               \\ \hline
% do you want to go to a movie tonight? & better out?                   \\ \hline

% \end{tabular}
% \caption{Example of predicted response sentences with given messages. The ``Chandler" language model was used for evaluation.}
% \end{table}

\section{Related Work}
\label{related work}
Researchers have proposed language models using RNN, which learns the probability of next sequence data at the character or word level \cite{sutskever2011generating,graves2013generating}. The proposed language models were tested on web corpora (i.e. Wikipedia, news articles) and qualitative examples showed their applicability. \cite{sutskever2014sequence} proposed a sequence-to-sequence learning algorithm with RNN and long short-term memory (LSTM) architecture \cite{hochreiter1997long}, and \cite{cho2014learning} proposed RNN encoder-decoder architecture. Those studies were applied to the machine translation problem.

% With large amounts of data, it learned the matching probability between different languages and predicted the translated sequence. A language sentence is encoded as an n-dimensional context vector, and the vector is used for predicting the sequence of the target language.

Recently, the RNN machine translation approach was extended to the short message generation problem \cite{sordoni2015NAACL-HLT}. Considering the message and response as a translation problem, the Neural Responding Machine achieved 40\% accuracy for both contextually and syntactically proper response generations with twitter-like micro-blogging data \cite{shang2015neural}. Those studies were similar to our research in the sense that both target message-reply prediction language model using RNN. Our research, however, differs in that it updates a general language model to a personalized language model with user data separately, whereas the previous research trained a language model with the data, as a whole, in same place.

% Recently, the RNN machine translation approach was extended to the short message generation problem. Considering the message and response as a translation problem, the Neural Responding Machine achieved 40\% accuracy for both contextually and syntactically proper response generation with twitter-like micro-blogging data \cite{shang2015neural}. Similarly, there was an attempt to keep the dialogue context for long terms. The neural conversational model embedded the dialogue and tried to keep the conversational context \cite{sordoni2015NAACL-HLT}. Those studies were similar to our research in the sense that both target message-reply prediction language model using RNN. Our research, however, differs in that it updates a general language model to a personalized language model with user data separately, whereas the previous research trained a language model with the data, as a whole, in same place.

In the commercial sphere, Google recently released a smart-reply service that could generate a response to a given email by using a sequence-to-sequence learning model  \cite{GoogleSR}. There was another trial on the generation of  responses in technical troubleshooting discourses \cite{vinyals2015neural}. This research also required complete data in one place and did not provide a personalized model.

Moreover, many researchers have conducted studies on transfer learning. \cite{bengio2011deep,bengio2012deep} suggested that a base-trained model with general data could be transferred to another domain. Recently, \cite{yosinski2014transferable} showed, through experiments, that the lower layers tended to have general features whereas the higher layer tended to have specific features. However, none of this research was applied to an RNN language model.

To adapt a neural network model to an embedded system with limited resources, \cite{kim2015compression} \cite{han2015learning} reduced the size of the model by pruning the unnecessary connections within it. It repeatedly tried to reduce the model size without accuracy degradation. This research inspired us to a considerable extent. It applied a neural model to mobile devices. However, the research focused on reducing the model size using a powerful machine and releasing the final model to an embedded system, whereas ours investigated how to train a model within mobile devices so that private user data could be kept.

\section{Conclusion}
\label{conclusion}

% We propose a efficient method for training a personalized model using the LSTM-RNN model. To preserve users' privacy, we suggest various transfer learning schemes so that the personalized language model can be generated within the user's local environment. The proposed schemes ``surplus layer' and ``fixed-n layer' shows higher generalization performance whereas it trains only reduced number of parameters than baseline model. Furthermore, we propose a novel metric to measure the similarity between a user style and that of the personalized language model output. With this metric, we evaluate the personalized language model and found a lower value between the output of the personalized language model and the same user dialogue than that between the output of the personalized language model and other user dialogue. The qualitative test result also indicates that the output of the model is similar to that of the user's style.

We propose an efficient method for training a personalized model using the LSTM-RNN model. To preserve users' privacy, we suggest various transfer learning schemes so that the personalized language model can be generated within the user's local environment. The proposed schemes ``surplus layer' and ``fixed-n layer' shows higher generalization performance whereas it trains only reduced number of parameters than baseline model. The quantitative and qualitative test result indicate that the output of the model is similar to that of the user's style.

It is certain that our proposed method reveals the applicability of the RNN-based language model in a user device with the preservation of privacy. Furthermore, with our method the personalized language model can be generated with a smaller amount of user data than the huge amount of training data that is usually required in the traditional deep neural network discipline. In the future work, we aim to visualize the deep neural network and to investigate the specific relationship among users' language styles and the LSTM cells in the network. This approach seems likely to uncover enhanced learning schemes that require less data than was previously necessary.

\bibliography{aaai}
\bibliographystyle{aaai}

\end{document}